\documentclass{article}
\usepackage{ijcai25}

\usepackage{times}
\usepackage{soul}
\usepackage{url}
\usepackage{amsmath,amssymb,amsfonts}
\usepackage[hidelinks]{hyperref}
\usepackage[utf8]{inputenc}
\usepackage[small]{caption}
\usepackage{graphicx}
\usepackage{amsmath}
\usepackage{amsthm}
\usepackage{booktabs}
\usepackage{multirow}
\usepackage{subcaption}
\usepackage[ruled]{algorithm2e}
\usepackage{algorithmic}
\usepackage[switch]{lineno}

\title{RotateKV: Accurate and Robust 2-Bit KV Cache Quantization\\for LLMs via Outlier-Aware Adaptive Rotations }
\author{
Zunhai Su\textsuperscript{1},
Zhe Chen\textsuperscript{2},
Wang Shen\textsuperscript{2}, 
Hanyu Wei\textsuperscript{1},
Linge Li\textsuperscript{2}, 
Huangqi Yu\textsuperscript{2},
Kehong Yuan\textsuperscript{1}\thanks{Corresponding author.}
\\
 \textsuperscript{1}Tsinghua University, \textsuperscript{2}Huawei Technologies Co., Ltd
\\
\{zh-su23,wei-hy23\}@mails.tsinghua.edu.cn \\
\{chenzhe49,shenwang1,lilinge,yuhuangqi\}@huawei.com,
yuankh@sz.tsinghua.edu.cn
}
\begin{document}
\maketitle
\begin{abstract}
Key-Value (KV) cache facilitates efficient large language models (LLMs) inference by avoiding recomputation of past KVs. 
As the batch size and context length increase, the oversized KV caches become a significant memory bottleneck, highlighting the need for efficient compression.
Existing KV quantization rely on fine-grained quantization or the retention of a significant portion of high bit-widths caches, both of which compromise compression ratio and often fail to maintain robustness at extremely low average bit-widths.
In this work, we explore the potential of rotation technique for 2-bit KV quantization and propose RotateKV, which achieves accurate and robust performance through the following innovations:
(i) Outlier-Aware Rotation, which utilizes channel-reordering to adapt the rotations to varying channel-wise outlier distributions without sacrificing the computational efficiency of the fast Walsh-Hadamard transform (FWHT);
(ii) Pre-RoPE Grouped-Head Rotation, which mitigates the impact of rotary position embedding (RoPE) on proposed outlier-aware rotation and further smooths outliers across heads;
(iii) Attention-Sink-Aware Quantization, which leverages the massive activations to precisely identify and protect attention sinks.
RotateKV achieves less than 0.3 perplexity (PPL) degradation with 2-bit quantization on WikiText-2 using LLaMA-2-13B, maintains strong CoT reasoning and long-context capabilities, with less than 1.7\% degradation on GSM8K, outperforming existing methods even at lower average bit-widths.
RotateKV also showcases a 3.97× reduction in peak memory usage, supports 5.75× larger batch sizes, and achieves a 2.32× speedup in decoding stage.
\end{abstract}
\begin{figure}[t]
    \centering
\includegraphics[width=1\linewidth]{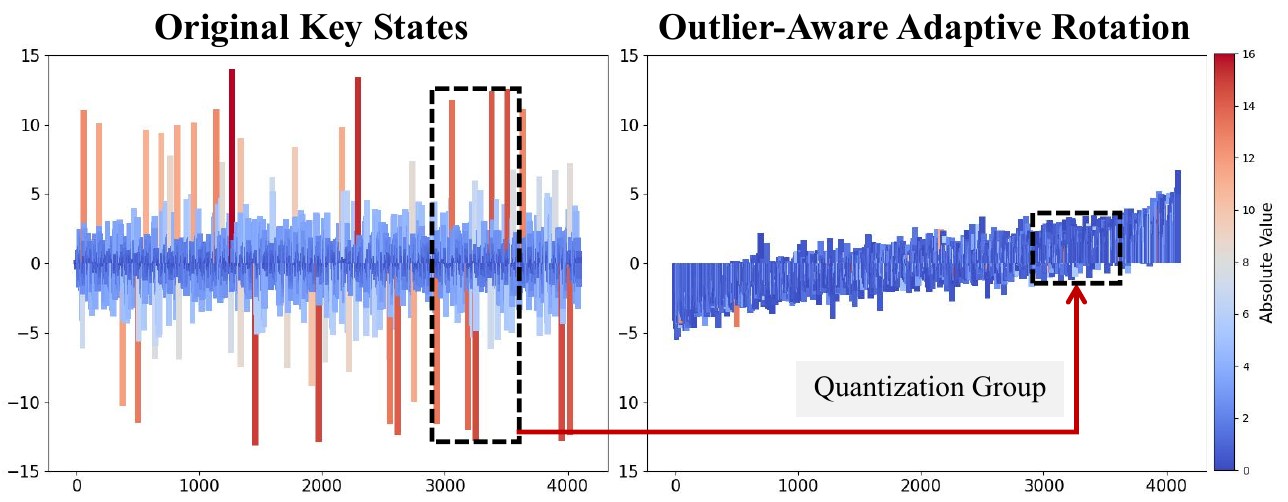}
\vspace{-7mm}
    \caption{The distribution of Keys in Layer 10 of LLaMA-2-7B. 
    The proposed outlier-aware adaptive rotation demonstrates outstanding capability in reducing outliers. }
    \label{fig:main image}
\vspace{-5mm}
\end{figure}
\begin{figure*}[t]
\vspace{-8mm}
    \centering
    % Subfigure 1
    \begin{subfigure}[b]{0.23\linewidth}
        \centering \includegraphics[width=\linewidth]{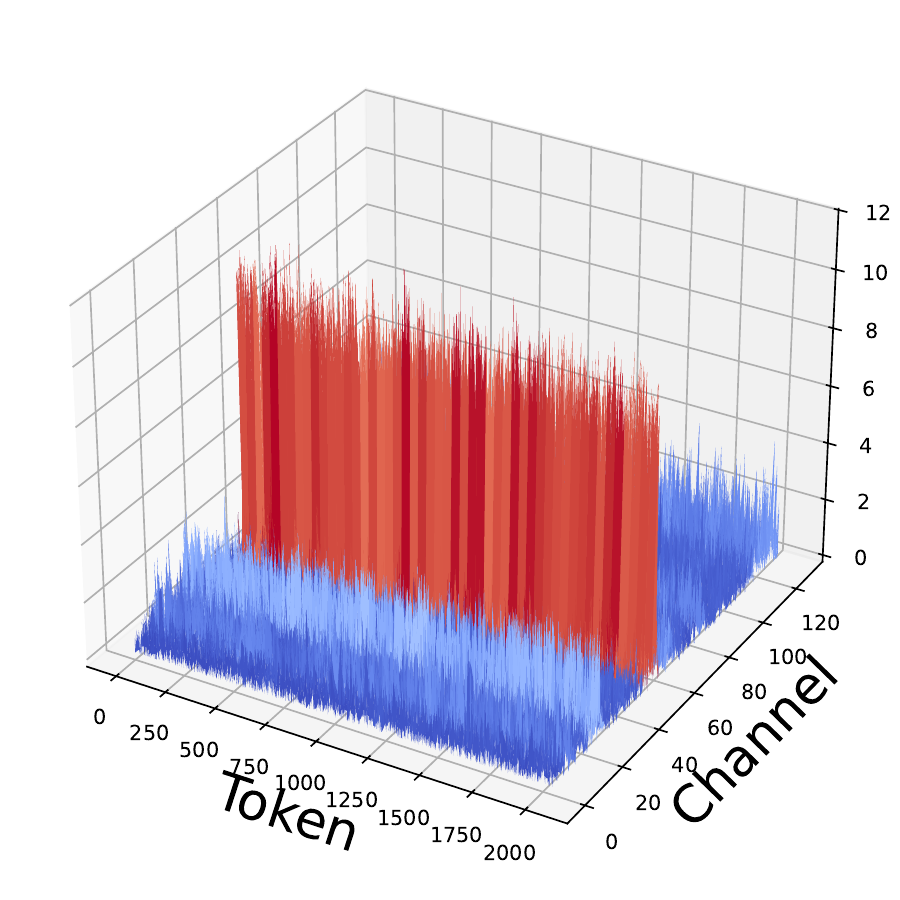}    
        \vspace{-5mm}
        \caption{Layer 10 Head 1}
    \end{subfigure}
    \begin{subfigure}[b]{0.23\linewidth}
        \centering      \includegraphics[width=\linewidth]{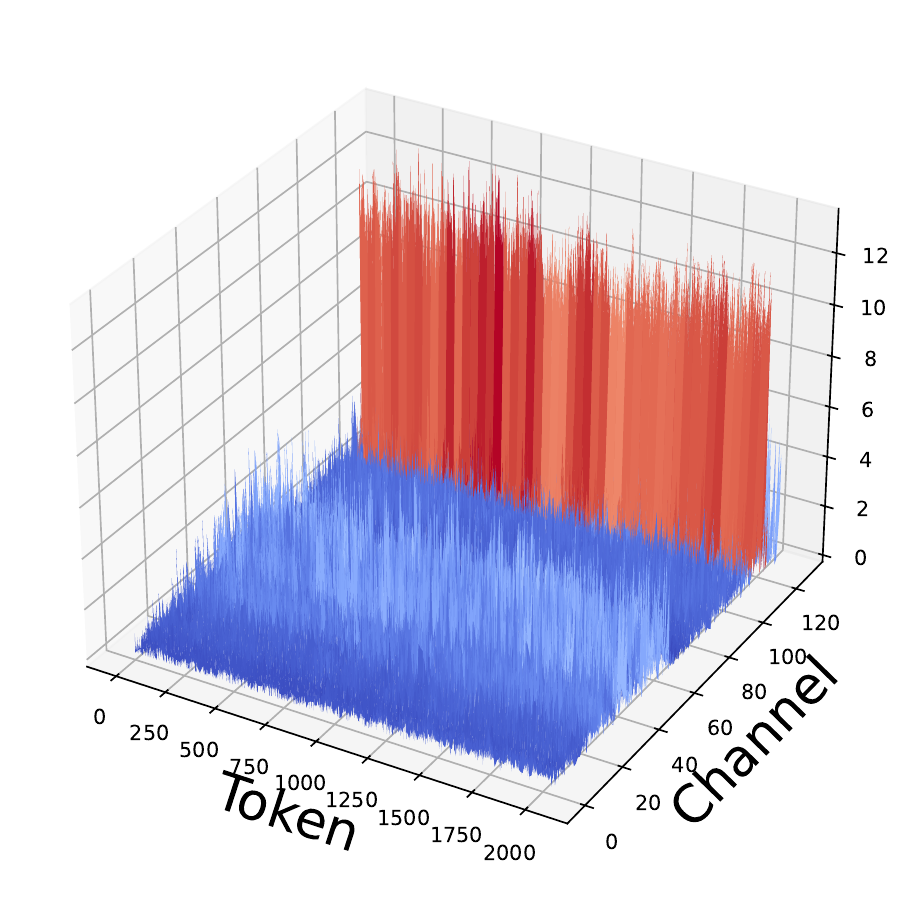} 
        \vspace{-5mm}                
        \caption{Layer 10 Head 2}
    \end{subfigure}
    \begin{subfigure}[b]{0.23\linewidth}
        \centering
    \includegraphics[width=\linewidth]{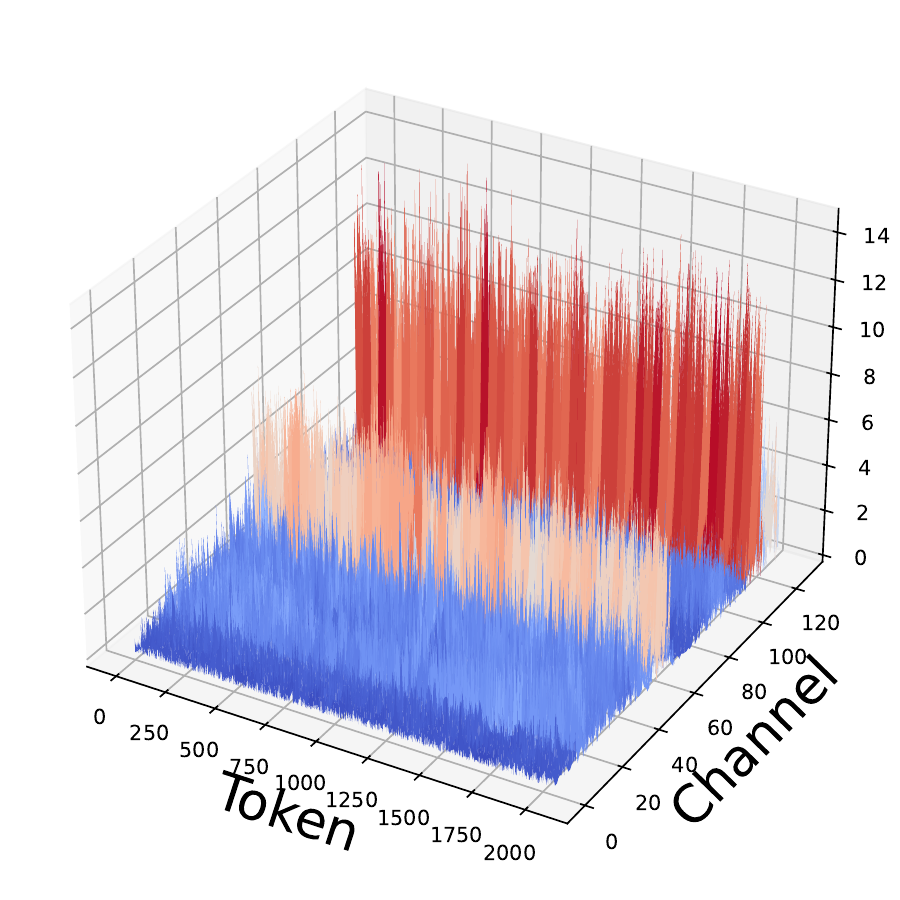}   
            \vspace{-5mm}
        \caption{Layer 10 Head 30}
    \end{subfigure}
    \begin{subfigure}[b]{0.23\linewidth}
        \centering
        \includegraphics[width=\linewidth]{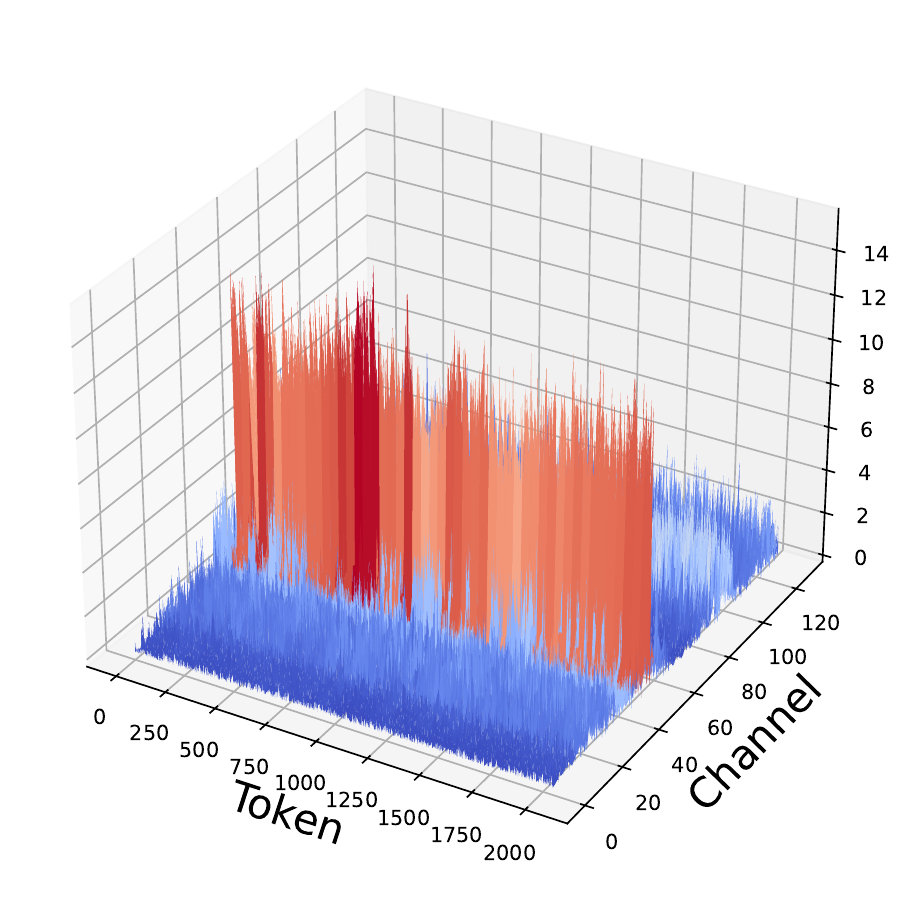} 
                \vspace{-5mm}
        \caption{Layer 10 Head 31}
    \end{subfigure}
            \vspace{-2mm}
    \caption{Magnitude of of LLaMA-2-7B Keys.
A small number of channels exhibit disproportionately large magnitudes, and these outlier channels vary across different attention heads.
    }
        \label{channel-wise}
\vspace{-5mm}
\end{figure*}
\vspace{-5mm}
\section{Introduction}
Large language models (LLMs) have attracted considerable attention due to their remarkable capabilities in next-token-prediction generation tasks \cite{zhao2023survey}. 
A critical technique of LLMs is the Key-Value (KV) cache, which avoids recomputation by caching the KVs generated by the attention layer in each Transformer block \cite{vaswani2017attention}. 
However, as the batch size and context length increase, the growing size of KV caches not only leads to significant memory consumption, but also renders LLM inference memory-bound, limiting system throughput \cite{liu2024kivi,kang2024gear}.
Low-bit quantization is widely used to compress LLMs for enhanced memory and time efficiency, covering various aspects, including weight-only quantization \cite{frantar2022gptq,lin2024awq}, weight-activation quantization \cite{xiao2023smoothquant,ashkboos2024quarot,liu2024spinquant,saxena2024resq}, and KV cache quantization.
Existing studies on KV cache quantization often employ techniques like fine-grained per-channel quantization \cite{liu2024kivi,kang2024gear} or mixed-precision quantization \cite{he2024zipcache,yang2024no,duanmu2024skvq,hooper2024KVQuant}, which retain a significant portion of high bit-widths caches.
As a result, these methods compromise compression efficiency and fail to maintain robustness under high compression ratio or extremely low average bit-widths.

Recently, Hadamard-transform-based rotation technique has demonstrated significant effectiveness in mitigating outliers in 4-bit LLM quantization, with studies such as \cite{ashkboos2024quarot,liu2024spinquant,saxena2024resq} leveraging rotation technique to achieve 4-bit quantization of weights, activations and KV caches.
However, the potential of rotation methods for extremely low-bit KV cache quantization has yet to be fully explored.
In this study, we thoroughly analyze the limitations of existing KV rotation settings and propose RotateKV, a tuning-free method that ensures superior outlier management through the following innovations (see Figure \ref{fig:main image} for an illustrative example):

\indent\textbullet\ \textbf{Outlier-Aware Rotation.}
Recent research \cite{liu2024kivi,hooper2024KVQuant} on outlier behavior in KV cache highlight the Keys contain a small number of channels with disproportionately large magnitudes, while these outlier channels vary across different attention heads, as shown in Figure \ref{channel-wise}.
However, existing Key rotation rely on the fast Walsh-Hadamard transform (FWHT), where the same Hadamard matrix is applied to each head across all layers.
Although the FWHT reduces the overhead of online computation, its rotation fails to adapt to the varying distributions of channel-wise outliers.
Through a comprehensive comparative analysis of various outlier-aware strategies, we propose outlier-aware rotation, which effectively enhances the adaptability of rotation to diverse outlier distributions while maintaining the computational efficiency of FWHT.

\indent\textbullet\ \textbf{Pre-RoPE Grouped-Head Rotation.} 
Existing rotations of Keys are performed by applying the rotation to each head of the Query and Key after rotary position embedding (RoPE) \cite{su2024roformer}, which presents two significant limitations that reduce the effectiveness of outlier-aware rotation in mitigating outliers.
First, RoPE disrupts the channel magnitude consistency of the Key tensors. 
Second, applying the rotation through the attention calculation of each head confines outlier reduction to individual heads. 
To address these issues, we propose a novel pre-RoPE rotation pipeline, which reduces the impact of RoPE on rotation and enables grouped-head rotation to smooth outliers across multiple heads.

\indent\textbullet\ \textbf{Attention-Sink-Aware Quantization.} 
Retaining attention sinks \cite{xiao2023efficient} is crucial for maintaining model performance during compression \cite{duanmu2024skvq,hooper2024KVQuant,liu2024intactkv}. 
Existing methods primarily focus on retaining sink tokens at the beginning of sequences, overlooking those in other positions \cite{yu2024unveiling,sun2024massive}. 
Inspired by recent studies on the interpretability of attention sinks \cite{sun2024massive,guo2024active}, we propose attention-sink-aware quantization, which leverages the correlation between massive activations \cite{sun2024massive} and attention sinks to precisely identify and retain additional salient sink tokens.

With these innovations, RotateKV fully exploits the potential of rotation for extremely low-bit quantization, addressing the limitations of current KV quantization methods and delivering both accurate and robust 2-bit quantization.
\textbf{Our contributions are summarized as follows:}

\indent\textbullet\ We identify the limitation of existing Key rotation that applies the same rotation to all attention heads.
We propose outlier-aware rotation, which improves adaptability to diverse outlier distributions, while maintaining the computational efficiency of FWHT.

\indent\textbullet\ Building on our analysis of the impact and limitations of existing post-RoPE rotations, we propose pre-RoPE rotation that effectively mitigates RoPE's influence on rotations while unlocking the potential for grouped-head rotation.

\indent\textbullet\ We propose attention-sink-aware quantization, an innovative approach that provides new insights into identifying attention sinks and extends the current practice of retaining only the sink tokens at the beginning of the sequence.

\indent\textbullet\ To the best of our knowledge, RotateKV is the first method to fully explore and harness the potential of rotation in extremely low-bit KV quantization.
Extensive experiments demonstrate that RotateKV outperforms state-of-the-art methods while achieving higher compression ratio.
RotateKV achieves less than 0.3 perplexity (PPL) degradation with 2-bit on WikiText-2 \cite{merity2016pointer} using LLaMA-2-13B \cite{touvron2023llama}. 
Evaluations on the GSM8K \cite{cobbe2021gsm8k} demonstrates that RotateKV maintains strong chain-of-thought (CoT) reasoning capability, with less than 1.7\% performance degradation compared to FP16 baseline.
RotateKV also exhibits negligible performance degradation in the 40K context-length Needle-in-a-Haystack (NIAH) \cite{10.5555/3692070.3692634} test, as well as in eight challenging long-context and multi-modal tasks chosen from LongBench \cite{bai2023longbench} and MileBench \cite{song2024milebench}.
With the current implementation, RotateKV can reduce peak memory usage by 3.97×, supports 5.75× larger batch sizes, and achieves a 2.32× decoding speedup.
\vspace{-2mm}
\section{Related Work}
\subsection{KV Cache Quantization}
Quantization reduces the bit-widths of numerical representation to compress the KV cache, effectively decreasing memory usage and alleviating the memory-bound challenge. 
Generally, existing KV cache quantization methods can be categorized into two types based on the quantization dimension of Keys: per-channel and per-token.
Due to significant outliers along the channel dimension in Keys, methods such as KIVI \cite{liu2024kivi}, Gear \cite{kang2024gear}, and KVQuant \cite{hooper2024KVQuant} adopt per-channel quantization to mitigate quantization errors. 
However, these approaches often require fine-grained quantization or the retention of a certain proportion of outliers unquantized to preserve model performance, both of which compromise the compression ratio.
On the other hand, due to the autoregressive nature of LLMs, where tokens are predicted sequentially, per-token quantization is commonly used in methods such as ZipCache \cite{he2024zipcache}, MiKV \cite{yang2024no}, SKVQ \cite{duanmu2024skvq}, and other less aggressive KV quantization methods \cite{ashkboos2024quarot,liu2024spinquant,shah2024flashattention}.
These approaches typically focus on the saliency differences between tokens, allocating relatively higher bit-widths to store a significant proportion of salient tokens. However, this also results in a compromised compression ratio.
Unlike these methods, RotateKV employs per-token quantization but eliminates the need to store large proportions of high-bit caches. 
Besides, experiments show that RotateKV offers superior outlier management compared to per-channel approaches, ensuring robust performance even at higher compression ratio.
\vspace{-2mm}
\section{Preliminary}
\begin{figure*}[t]
\vspace{-5mm}
\centering
  \includegraphics[width=0.9\textwidth]{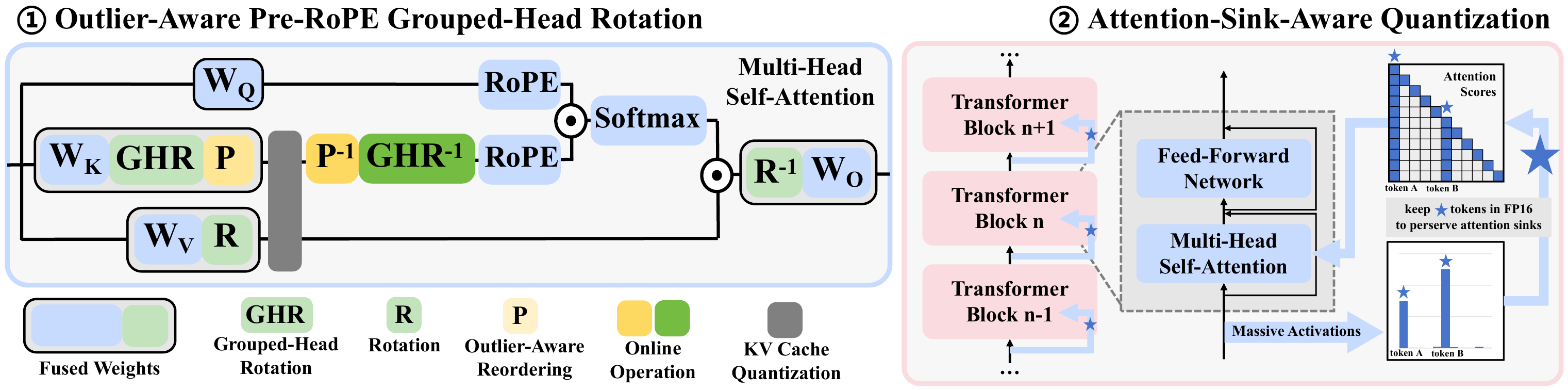}
  \vspace{-3mm}
      \caption{Overview of RotateKV.
      On the left is the outlier-aware rotation combined with the pre-RoPE pipeline.
On the right, we demonstrate attention-sink-aware quantization. 
Since attention is concentrated on the massive activations, we can identify attention sinks in the current attention layer by utilizing the token indices of these massive activations from the output of the previous decoder block.}
    \label{fig:overview}
  \vspace{-3mm}
\end{figure*}
\subsection{Inference Process of LLMs}
The LLM inference process consists of two stages: the prefill phase and the decoding phase.

\textbf{Prefill Phase.} 
The model processes the token sequence generated from the prompt and generates the initial output token, with each attention layer computing and caching KV pairs. 
Let $\mathbf{X}\in{\mathbb{R}^{l_{prompt}\times d}}$ represent the input embeddings, where $l_{prompt}$ is the length of the input token sequence and $d$ is the model's hidden size. 
In each attention layer, the KV cache can be derived as follows:
\begin{equation}
    K = \mathbf{X} \cdot W_k,  V = \mathbf{X} \cdot W_v,
\end{equation}
\begin{equation}
\mathbf{K_{cache}} \leftarrow K, \mathbf{V_{cache}} \leftarrow V,
\end{equation}
where $W_k, W_v\in\mathbb{R}^{d\times d}$ are the weight matrices for the Key and Value calculations, respectively.

\textbf{Decoding Phase.} 
The model takes a single token as input.
Let $\mathbf{t}\in{\mathbb{R}^{1\times d}}$ as the input embedding. 
Each attention layer computes \( t_Q \), \( t_K \) and \( t_V \) as follows:
\begin{equation}
    t_Q = \mathbf{t} \cdot W_q, t_K = \mathbf{t} \cdot W_k, t_V = \mathbf{t} \cdot W_v.
\end{equation}
Then, \( t_K \) and \( t_V \) are employed to update the KV cache, with the complete KV cache supporting subsequent computations.
\begin{equation}
    K \leftarrow \mathrm{concat}(\mathbf{K_{cache}}, t_K),
    V \leftarrow \mathrm{concat}(\mathbf{V_{cache}}, t_V),
\end{equation}
\begin{equation}
    t_O = \mathrm{softmax}(t_Q \cdot K^\top \cdot d^{-\frac{1}{2}})\cdot V.\\
\end{equation}
\subsection{Quantization}
\label{quantization}
The n-bit asymmetric integer quantization and dequantization processes can be expressed as:
\begin{equation}
Q(X) = clamp\left(\left\lfloor \frac{X}{{scale}} \right\rceil + zero, 0, 2^n - 1 \right),
\end{equation}
\begin{equation}
X' = scale \cdot (Q(X) - zero),
\end{equation}
\begin{equation}
    scale = \frac{clipped\_max(X) - clipped\_min(X)}{2^n - 1} ,
\end{equation}
\begin{equation}
    zero = - \left\lfloor \frac{clipped\_min(X)}{scale} \right\rceil,
\end{equation}
where $\left\lfloor\cdot\right\rceil$ indicates round operation. 
$Q(X)$ and $X'$ denote the quantized and dequantized values of $X$. 
The $clamp$ operation constrains the values within a specified range. 
$clipped\_max(X)$ and $clipped\_min(X)$ denote the operations that truncate the maximum and minimum values of $X$. 
\subsection{Hadamard Matrix}
Hadamard Matrix $R$ is a specific type of orthogonal matrix characterized by entries proportional to $\{ +1, -1 \}$. 
The Hadamard matrix also satisfies the definition of a rotation matrix, as it is an orthogonal matrix with $\det(R)=1$. 
QuIP \cite{chee2024quip} demonstrates that multiplying an rotation
matrix leads to a reduction in the maximum entry relative to its norm, effectively mitigating outliers and enhancing quantizability.
Walsh-Hadamard Matrix is a particular instance of Hadamard matrix generated recursively as follows, the subscript denoting the dimension of matrix, where \( k \in \mathbb{Z}^+ \):
\begin{equation}
    H_1 = \begin{bmatrix} 1 \end{bmatrix}, \quad
    H_{2^k} = \frac{1}{\sqrt{2}} \begin{bmatrix}
    H_{2^{(k-1)}} & H_{2^{(k-1)}} \\
    H_{2^{(k-1)}} & -H_{2^{(k-1)}}
    \end{bmatrix}.
\label{walsh-hadamard}
\end{equation}
The scaling factor \( \frac{1}{\sqrt{2}} \) ensures normalization. 
The recursive structure of the Walsh-Hadamard matrix enables efficient computation via the FWHT algorithm \cite{hedayat1978hadamard}, reducing computational complexity of matrix-vector multiplication to \( O(n \log n) \), where $n$ represents the dimension of the matrix. 
\vspace{-2mm}
\section{Methodology}
Section~\ref{Channel-Aware Rotation} outlines the limitations of existing outlier-unaware Key rotation and proposes outlier-aware rotation.
In Section~\ref{Pre-RoPE Grouped-Head Rotation}, we analyze the impact of RoPE on rotation, then propose pre-RoPE grouped-head rotation.
Section~\ref{Attention-Sink-Aware Quantization} further introduces attention-sink-aware quantization.
Section~\ref{summary} provides a comprehensive summary of RotateKV.
The overview of RotateKV is illustrated in Figure \ref{fig:overview}.
\subsection{Outlier-Aware Rotation}
\label{Channel-Aware Rotation}
\subsubsection{Existing Outlier-Unaware Key Rotation}
\begin{figure}[t]
    \vspace{-3mm}
        \centering
\includegraphics[width=0.9\linewidth]{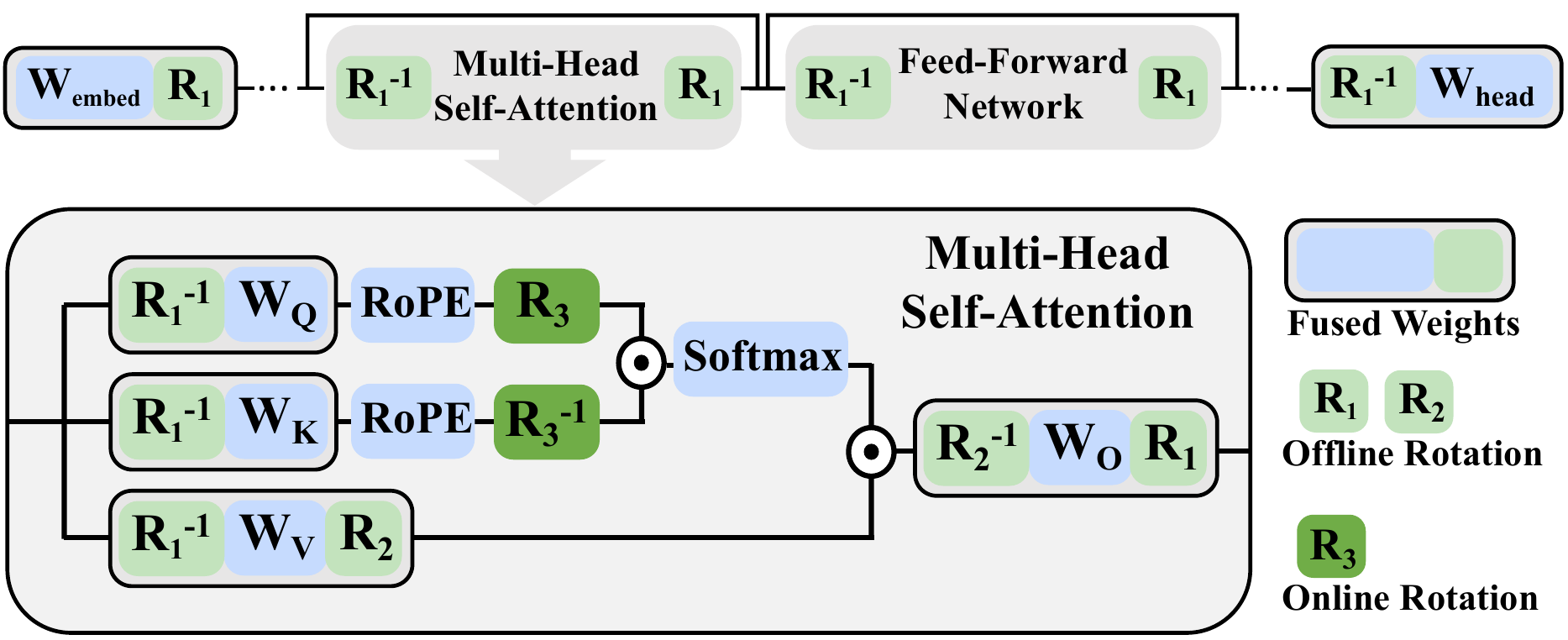}
    \vspace{-3mm}
        \caption{Existing rotation paradigm.}
        \label{fig:4-1}
    \vspace{-3mm}
    \label{spinquant}
        \vspace{-1mm}
\end{figure}

When applying quantization, outliers pose a persistent challenge as they expand the quantization range, reducing the effective bits available for most values. 
Recently, LLM quantization research \cite{chee2024quip,ashkboos2024quarot,liu2024spinquant,saxena2024resq,lin2024duquant,shah2024flashattention} has focused on Hadamard-transform-based rotation technique, which involves multiplying a Hadamard matrix to reduce outliers and improve quantizability.
As shown in Figure \ref{spinquant}, existing rotations within the attention layer can be classified into two categories: offline rotations (e.g., \(R_1\) and \(R_2\)), which can be fused into the weights prior to inference to eliminate the overhead of online computations, and online rotations (e.g., \(R_3\)), which are performed dynamically using the FWHT.
This differentiation exists because the Query and Key computation relies on RoPE, rendering it incompatible with offline rotations. 
To reduce the computational overhead during inference, Key rotations typically utilize the efficient FWHT algorithm.
For example, QuaRot \cite{ashkboos2024quarot} leverages offline random Hadamard matrices and online FWHT to achieve outlier-free 4-bit LLM inference. 
SpinQuant \cite{liu2024spinquant} employs Cayley optimization to enhance offline rotation but continues to utilize the FWHT for online rotations.
Due to the structure of the Walsh-Hadamard matrix, as shown in Equation \ref{walsh-hadamard}, the rotations for each attention head with the same dimension rely on the same Walsh-Hadamard matrix.
This constraint limits adaptability to varying channel-wise outlier distributions.
\subsubsection{Outlier-Aware Rotation}
While adjusting rotation for pre-rotation distributions seems straightforward, it is impractical. 
Modifying the Hadamard matrices invalidates the FWHT algorithm, increases computational overhead, and necessitates storing multiple matrices, thereby further reducing compression ratio.
To adapt rotations to varying channel-wise outlier distributions without compromising the efficiency of FWHT, we introduce the outlier-aware rotation, which preserves the FWHT while enhancing it with outlier-aware operation derived from lightweight calibration. 
We conduct experiments using various enhancement strategies, including smoothing and reordering.
Smoothing \cite{xiao2023smoothquant,lin2024qserve} scales down the Keys by 
per-channel factors \(\lambda\) and correspondingly scaling up the associated channels in the Query:
\begin{equation}
    \mathbf{Q} \cdot \mathbf{K}^T = (\mathbf{Q} \mathbf{\Lambda}) \cdot (\mathbf{K} \mathbf{\Lambda}^{-1})^T, \quad \mathbf{\Lambda} = \text{diag}(\lambda).
\end{equation}
Reordering arranges channels across all heads to reduce outliers within each quantization group.
As shown in Table~\ref{tab:smooth}, experiments on different outlier-aware strategies demonstrate that smoothing fails to maintain performance at 2-bit quantization.
For reordering, we find that relying on clustering or other complex permutation methods is unnecessary. 
Instead, reordering the channels of each token using the indices sorted by the values after rotation can effectively reduce quantization errors (Table \ref{tab:quantizeerror}) and improve PPL.
Notably, the channel reordering indices are derived through fast calibration and remain consistent across all tokens throughout inference.
% We hypothesize that reordering after rotation operates by rearranging the columns of the Hadamard matrix in accordance with the distribution of outliers across channels.
The calibration procedure is detailed in Algorithm \ref{alg:calibration}.

% Figure illustrates the distributions and quantization error of Keys after applying calibration-based post-rotation reordering validating the improvements achieved through reordering.
% Intuitively, rotation mitigates magnitude disparities within quantization groups, while reordering after rotation capitalizes on the smoother distributions to further reduce quantization errors.
% Theoretically, reordering after rotation functions by rearranging the columns of the Hadamard matrix based on outlier distribuation, thereby enhancing the mitigation of outliers during rotation.
\begin{table}[t]
\centering
\vspace{-2mm}
\resizebox{0.6\columnwidth}{!}{%
\begin{tabular}{@{}l|cccc@{}}
\toprule
\multirow{2}{*}{Methods} & \multicolumn{4}{c}{LLaMA-2-7B PPL $\downarrow$} \\ \cmidrule(l){2-5} 
 & \multicolumn{1}{c|}{16bit} & 4bit & 3bit & 2bit \\ \midrule
Rotate-Only & \multicolumn{1}{c|}{\multirow{7}{*}{5.47}} & 5.62 & 6.16 & 8.05 \\
Smooth-Only & \multicolumn{1}{c|}{} & 5.71 & 6.65 & 16.97 \\
Reorder-Only & \multicolumn{1}{c|}{} & 5.73 & 6.47 & 8.26 \\ \cmidrule(r){1-1} \cmidrule(l){3-5} 
Rotate + Smooth & \multicolumn{1}{c|}{} & 5.62 & 6.13 & 6.93 \\
Smooth + Rotate & \multicolumn{1}{c|}{} & 5.61 & 6.02 & 6.83 \\ \cmidrule(r){1-1} \cmidrule(l){3-5} 
\textbf{Rotate + Reorder} & \multicolumn{1}{c|}{} & \textbf{5.63} & \textbf{6.07} & \textbf{6.33} \\
Reorder + Rotate & \multicolumn{1}{c|}{} & 7.43 & 36.94 & 231.43 \\ \bottomrule
\end{tabular}%
}
\vspace{-2mm}
\caption{Experiments on different outlier-aware strategies. 
We evaluate PPL with a sequence length of 2048 using the LLaMA-2-7B model on the WikiText-2 dataset.}
    \vspace{-3mm}
\label{tab:smooth}
\end{table}
\begin{algorithm}[t]
    \caption{Calibration for Reordering Indices}
    \label{alg:calibration}
    \textbf{Input}: Key states after rotation: $K \in \mathbb{R}^{b \times h \times s \times d}$ \\
    \textbf{Parameter}: Number of layers: $L$, batch size: $b$, number of attention heads: $h$, sequence length: $s$, dimensionality of each head: $d$. \\
    \textbf{Output}: Reordering indices $P_l$ for each layer $l \in \{1, 2, \dots, L\}$\\
    \begin{algorithmic}[1]
        \FOR{$l = 1$ \textbf{to} $L$}
            \STATE Reshape $K \in \mathbb{R}^{b s \times h d}$.
\STATE Compute $channel\_sum^l = \sum_{i=1}^{b s} K[i, :]$.
      \STATE Sort indices: $P_l = \text{argsort}(channel\_sum^l)$.
        \ENDFOR
        \STATE \textbf{return} $P_l$ for each $l \in \{1, 2, \dots, L\}$.
    \end{algorithmic}
\end{algorithm}
\begin{figure*}[t]
\vspace{-5mm}
\centering
\begin{subfigure}[b]{0.28\textwidth}
    \centering
    \vspace{-10mm}
    \includegraphics[width=\textwidth]{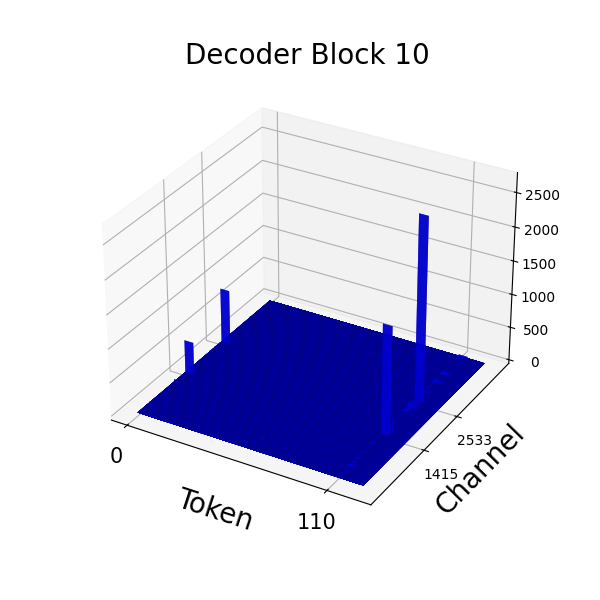}
    % \vspace{mm}
    \caption{Outputs of Decoder Block 10.}
    \label{fig:mass_a}
\end{subfigure}\hfill
\begin{subfigure}[b]{0.7\textwidth}
    \centering
    \includegraphics[width=\textwidth]{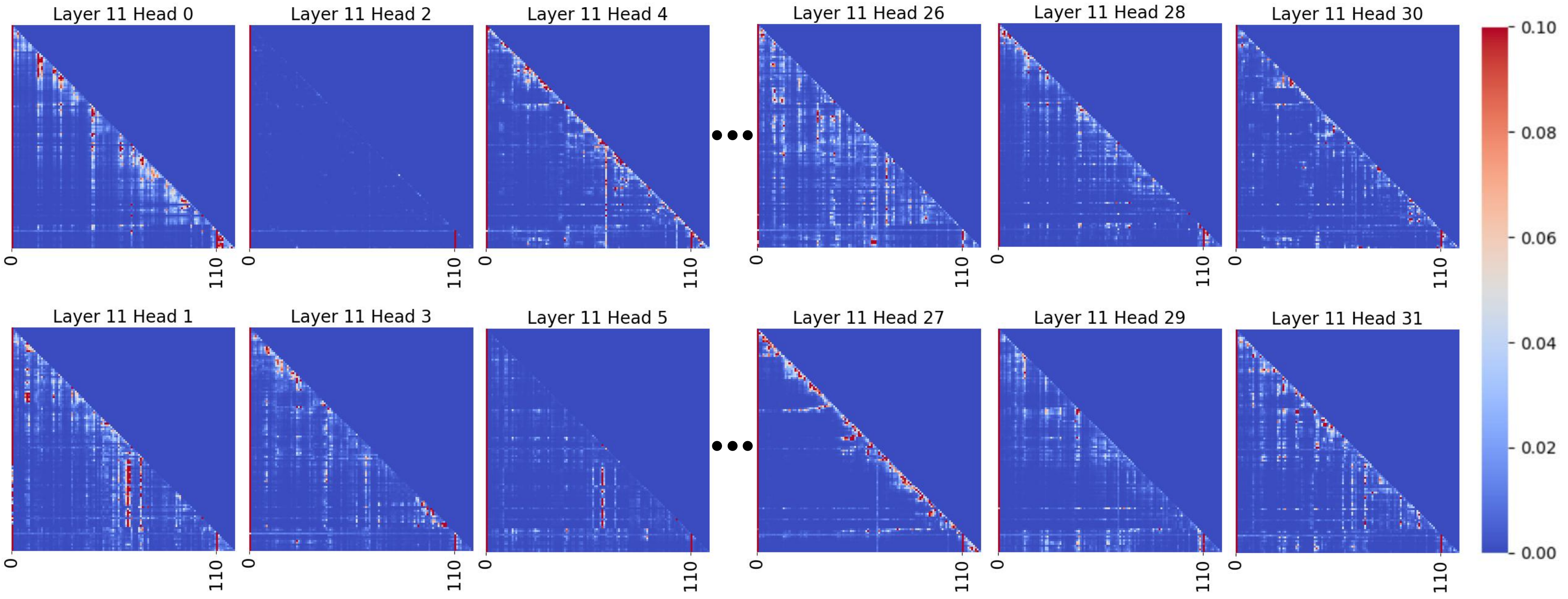}
    \vspace{-5mm}
        \caption{In this example, attention sinks are observed at tokens 0 and 110.}
    \label{fig:mass_b}
\end{subfigure}
    \vspace{-2mm}
\caption{Visualizations of Decoder Block 10 outputs and the attention scores in Attention Layer 11 from the LLaMA-2-7B using input from WikiText-2 dataset. 
As shown in Figure \ref{fig:mass_a}, massive activations occur at tokens 0 and 110, in channels 1415 and 2533. 
In the subsequent attention layer, attention is focused on tokens 1 and 110 across all heads, as illustrated in Figure \ref{fig:mass_b}.}
\label{fig:massive}
    \vspace{-5mm}
\end{figure*}
\begin{figure}[t]
    \centering
    \vspace{-2mm}
    % First row: two images side by side using subfigures
    \begin{subfigure}[b]{0.45\linewidth}
        \centering
        \includegraphics[width=\linewidth]{new_pdf/Layer10Head31KeyStates.pdf}
        \vspace{-7mm}
        \caption{Key states before RoPE.}
        \label{fig:image1}
    \end{subfigure} \hfill
    \begin{subfigure}[b]{0.45\linewidth}
        \centering
        \includegraphics[width=\linewidth]{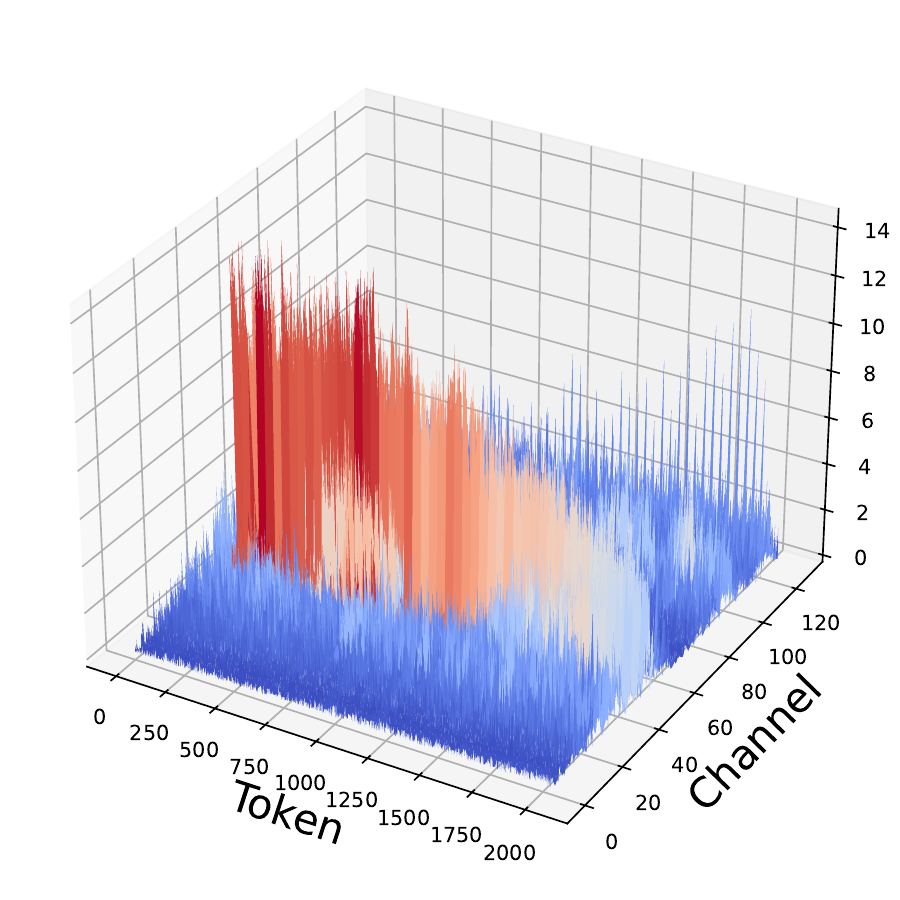}
        \vspace{-7mm}
        \caption{Key states after RoPE.}
        \label{fig:image2}
    \end{subfigure}

    % Second row: a single image with a         \vspace{-5mm}
    \begin{subfigure}[b]{0.8\linewidth}
        \centering
        \includegraphics[width=\linewidth]{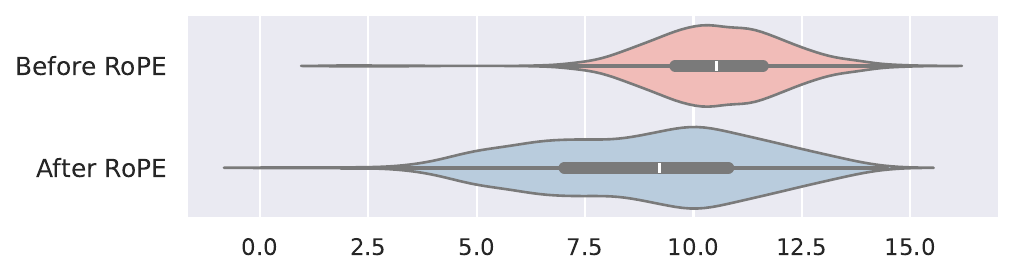}
        \vspace{-5mm}
        \caption{Distribution of Channel 51 before and after RoPE.}
        \label{fig:image3}
    \end{subfigure}
            \vspace{-2mm}
    \caption{Figures \ref{fig:image1} and \ref{fig:image2} illustrate the Key of LLaMA-2-7B, Layer 10, Head 31, before and after RoPE respectively. Figure \ref{fig:image3} presents the distribution of the outlier channel.}
                \vspace{-3mm}
    \label{fig:rope}
\end{figure}
\subsection{Pre-RoPE Grouped-Head Rotation}
\label{Pre-RoPE Grouped-Head Rotation}
\subsubsection{Existing Post-RoPE Key Rotation}
Recent LLMs, including LLaMA \cite{touvron2023llama}, Mistral \cite{jiang2023mistral} and Qwen \cite{bai2023qwen} use RoPE \cite{su2024roformer} to encode token positions. 
% For clarity, RoPE applied to a vector at position \( m \) can be expressed as:
% \begin{equation}
% \text{RoPE}(Q_m) = Q_m \cdot R_m,
% \end{equation}
% where \( Q_m \) represents the Query at position \( m \), and \( R_m \) denotes the corresponding transformation.
% The calculation of attention weights with RoPE can be simplified as follows:
% \begin{equation}
% Q R_m \cdot (K R_n)^T = Q R_{m-n} K^T,
% \end{equation}
% where \( R_{m-n} \) encodes relative positional relations. 
The positional encoding introduced by RoPE is applied to pairs of channels, resulting in a reduction of magnitude consistency within individual channel, as shown in Figure \ref{fig:rope}. 
Previous work \cite{hooper2024KVQuant} also highlights that RoPE affects per-channel quantization performance. 
In this study, we observe that the inconsistency in the magnitudes of channel-wise outliers significantly undermines the effectiveness of outlier-aware rotation. 
As demonstrated in Table \ref{tab:quantizeerror}, the quantization error increases by 145\% after applying RoPE.
% Additionally, as shown in Figure \ref{spinquant}, the existing post-RoPE Key rotations, applied through the attention calculation of each head, restrict outlier reduction to individual heads.
\begin{table}[t]
% tobedone
\centering
\resizebox{0.5\columnwidth}{!}{%
\begin{tabular}{@{}lcc@{}}
\toprule
\multirow{2}{*}{Methods} & \multicolumn{2}{c}{MSE $\downarrow$} \\ \cmidrule(l){2-3} 
 & Pre-RoPE & Post-RoPE \\ \midrule
Rotation & 0.632 & 0.670 \\
\textbf{+ Reordering} & \textbf{0.247} & \textbf{0.605} \\ \midrule
Reordering & 0.477 & 0.568 \\
+ Rotation & NaN & NaN \\ \bottomrule
\end{tabular}%
}
\vspace{-3mm}
\caption{We assess the reconstruction error of the Key in LLaMA-2-7B under 2-bit quantization, using mean squared error (MSE) as the metric. 
The results indicate that RoPE significantly affects the effectiveness of outlier-aware rotation.}
 \vspace{-5mm}
\label{tab:quantizeerror}
\end{table}
\begin{table}[t]
\centering
\resizebox{0.85\columnwidth}{!}{%
\begin{tabular}{@{}ccccc@{}}
\toprule
\begin{tabular}[c]{@{}c@{}}Heads\\ per Group\end{tabular} & \begin{tabular}[c]{@{}c@{}}Numbers of\\ Group\end{tabular} & \begin{tabular}[c]{@{}c@{}}Dimension of \\ Rotation Matrix\end{tabular} & \begin{tabular}[c]{@{}c@{}}FLOPs\\ /Layer\end{tabular} & \begin{tabular}[c]{@{}c@{}}LLaMA-2-7B\\ PPL $\downarrow$\end{tabular} \\ \midrule
1 & 32 & 128 & 28672 & 7.81 \\
2 & 16 & 256 & 32768 & 7.24 \\
\textbf{4} & \textbf{8} & \textbf{512} & \textbf{36864} & \textbf{6.99} \\
8 & 4 & 1024 & 40960 & 6.91 \\
16 & 2 & 2048 & 45056 & 6.98 \\
32 & 1 & 4096 & 49152 & 6.86 \\ \bottomrule
\end{tabular}%
}
\vspace{-2mm}
\caption{Experiments on group size of grouped-head rotation. 
The PPL is evaluated on the LLaMA-2-7B using WikiText-2 dataset. }
\vspace{-2mm}
\label{tab:groupedhead}
\end{table}
\subsubsection{Pre-RoPE Grouped-Head Rotation }
To address the issues associated with post-RoPE rotation, we introduce a novel pipeline that applies the outlier-aware rotation before RoPE.
As illustrated in Figure \ref{fig:overview}, this design not only eliminates RoPE's negative impact on rotation but also facilitates the incorporation of rotation and reordering operations into the weights, thus requiring only the inverse operation during inference.
The proposed pre-RoPE rotation also enables grouped-head combined rotation, allowing for more effective outlier reduction across heads. 
Although increasing the head numbers improves PPL, it also incurs higher computational costs. 
Therefore, it is crucial to balance computational overhead and performance gains when determining the group size. 
Typically, as shown in Table~\ref{tab:groupedhead}, a group size of 4 heads is considered a reasonable choice.
% \begin{algorithm}[t]
%  \SetKwProg{Procedure}{Procedure}{\string:}{end Procedure}
%  \SetKwInput{AttentionParam}{Parameter}
% \AttentionParam{$W_q, W_k, W_v, W_o \in \mathbb{R}^{d\times d}$}
% \Procedure{Decoding}{
%     \KwIn{$\mathbf{t}\in{\mathbb{R}^{l\times d}}$, $\mathbf{K_{cache}}$, $\mathbf{V_{cache}}$}
%     $t_Q = \mathbf{t}\cdot W_q$,
%     $t_K = \mathbf{t}\cdot W_k$,
%     $t_V = \mathbf{t}\cdot W_v$\\
%     $t_Q \leftarrow \mathrm{split\_head}(t_Q)$,
%     $t_K \leftarrow \mathrm{split\_head}(t_K)$,   
%     $t_V \leftarrow \mathrm{split\_head}(t_V)$\\
%     $K \leftarrow \mathrm{concat}(\mathrm{dequant}(\mathbf{K_{cache}}), t_K)$,
%     $V \leftarrow \mathrm{concat}(\mathrm{dequant}(\mathbf{V_{cache}}), t_V)$\\
%     $\mathbf{K_{cache}} \leftarrow \mathrm{concat}(\mathbf{K_{cache}}, \mathrm{quantize}(t_K))$\\
%     $\mathbf{V_{cache}} \leftarrow \mathrm{concat}(\mathbf{V_{cache}}, \mathrm{quantize}(t_V))$\\
%     $K \leftarrow \mathrm{reorder^{-1}}(K)$\\   
%     $K \leftarrow \mathrm{grouped\_head\_rotate}(K)$\\ 
%     $t_Q \leftarrow \mathrm{RoPE}(t_Q), K \leftarrow \mathrm{RoPE}(K)$\\
%     $t_O \leftarrow \mathrm{softmax}(t_Q \cdot K^\top \cdot d^{-\frac{1}{2}})\cdot V$\\
%     $t_O \leftarrow \mathrm{concat\_head}(t_O)$\\
%     $t_O \leftarrow t_O \cdot W_o$ \\
%     \Return {$t_O$}
% }
%   \caption{RotateKV}
%   \label{algo: pre-RotateKV}
% \end{algorithm}
\subsection{Attention-Sink-Aware Quantization}
\label{Attention-Sink-Aware Quantization}
The research \cite{xiao2023efficient} indicates that LLMs tend to treat the initial token as a 'sink', assigning it disproportionately high attention scores. 
Moreover, \cite{hooper2024KVQuant,duanmu2024skvq,liu2024intactkv} highlight that the KVs associated with sink tokens are sensitive to quantization, and retaining only the initial token in FP16 can effectively enhance quantization. 
However, more recent studies \cite{yu2024unveiling,sun2024massive} suggest that these few-in-number attention sinks can emerge not only at the initial token but also at various other positions, whereas existing practices fail to retain them precisely. 
One key reason is that efficient attention computation relies on kernels such as FlashAttention \cite{shah2024flashattention}, which directly output the attention results without exposing the intermediate attention scores.
This prevents the dynamic identification of additional attention sinks beyond those fixed at the initial token. 
Inspired by recent studies on the interpretability of attention sinks \cite{sun2024massive,guo2024active}, we propose attention-sink-aware quantization, which leverages massive activations to identify additional sink tokens without relying on attention scores, thereby precisely retaining them during the quantization process.
Research on massive activations \cite{sun2024massive}—those activations in the residual sums of Transformer block outputs with significantly larger magnitudes than others—suggests that attention is concentrated on these activations.
Specifically, when massive activations occur, the corresponding tokens attract concentrated attention, forming attention sinks.
Figure \ref{fig:massive} provides a real example from LLaMA-2-7B \cite{touvron2023llama}.
Therefore, by identifying the massive activations, additional sink tokens can be pinpointed.
The process of attention-sink-aware quantization is outlined in Figure \ref{fig:overview}.
% The algorithm of attention-sink-aware quantization is provided in Algorithm.
\subsection{Summary of RotateKV}
\label{summary}
In summary, RotateKV first performs fast calibration to obtain reordering indices, then integrates grouped-head rotation and channel reordering into the Key weights. 
These operations result in outlier-aware rotation of the Keys, making them more suitable for quantization.
After updating the KV cache, inverse online reordering and rotation are applied.
For the Value, we adopt a simple offline rotation as shown in Figure \ref{fig:overview}, since Values do not contain outliers like Keys.
% \vspace{-5mm}
% \vspace{-20mm}
\section{Experiments}
\begin{table}[t]
\vspace{-5mm}
\centering
\resizebox{0.65\columnwidth}{!}{%
\begin{tabular}{@{}c|cccc@{}}
\toprule
\multirow{2}{*}{Methods} & \multicolumn{4}{c}{WikiText-2 PPL $\downarrow$} \\ \cmidrule(l){2-5} 
 & \begin{tabular}[c]{@{}c@{}}LLaMA\\ 2-7B\end{tabular} & \begin{tabular}[c]{@{}c@{}}LLaMA\\ 2-13B\end{tabular} & \begin{tabular}[c]{@{}c@{}}LLaMA\\ 3-8B\end{tabular} & \begin{tabular}[c]{@{}c@{}}Mistral\\ 7B\end{tabular} \\ \midrule
FP16 & 5.12 & 4.57 & 5.75 & 4.91 \\ \midrule
QuaRot-4bit & 5.15 & 4.60 & 5.83 & 4.93 \\
KVQuant-4bit & 5.14 & 4.59 & 5.79 & 4.94 \\
\textbf{RotateKV-4bit} & \textbf{5.13} & \textbf{4.58} & \textbf{5.79} & \textbf{4.92} \\ \midrule
QuaRot-3bit & 5.33 & 4.72 & 6.21 & 5.05 \\
KVQuant-3bit & 5.20 & 4.65 & 5.95 & 4.99 \\
\textbf{RotateKV-3bit} & \textbf{5.20} & \textbf{4.64} & \textbf{5.94} & \textbf{4.97} \\ \midrule
QuaRot-2bit & 8.94 & 6.96 & 21.43 & 6.62 \\
KVQuant-2bit & 5.59 & 4.95 & 6.75 & 5.34 \\
\textbf{RotateKV-2bit} & \textbf{5.50} & \textbf{4.84} & \textbf{6.69} & \textbf{5.24} \\ \bottomrule
\end{tabular}
}
\vspace{-2mm}
\caption{PPL evaluations on WikiText-2 with a sequence length of 4096. 
For both QuaRot and our method, the quantization group size was set to 128. 
For KVQuant, we adopted the scheme of retaining 0.5\% of numerical outliers in full precision during quantization.}
\vspace{-2mm}
\label{tab:PPL}

\end{table}
\subsection{Experiment Settings}
\subsubsection{Models and Tasks.} 
To validate the robustness of our method, we evaluate RotateKV on a variety of challenging tasks using both LLMs and visual-language models (VLMs), including LLaMA-2-7B/13B \cite{touvron2023llama}, LLaVA-v1.5-7B/13B \cite{liu2024visual}, Mistral-7B \cite{jiang2023mistral}, LLaVA-v1.6-Mistral-7B \cite{liu2023improved}, LLaMA-3-8B \cite{dubey2024llama}, and LLaMA-2-7B-80K \cite{fu2024data}.
We begin by evaluating the PPL on the WikiText-2 dataset \cite{merity2016pointer}.
Then, evaluation on the GSM8k dataset with CoT prompting is conducted to assess RotateKV's capability in handling complex CoT reasoning task.
We further test long-context and multi-modal tasks accuracy across eight tasks from the LongBench \cite{bai2023longbench} and MileBench \cite{song2024milebench}. 
To assess RotateKV's performance with extremely long contexts, we also evaluate it on the 40K context-length Needle-in-a-Haystack (NIAH) test using the LLaMA-2-7B-80K \cite{fu2024data} model.
% The source code used to reproduce the results and generate the visualizations is provided in the supplementary materials.
The source code for reproducing the results and generating the visualizations is available at \href{https://github.com/ZunhaiSu/RotateKV}{https://github.com/ZunhaiSu/RotateKV}.
\subsubsection{Quantization Settings.} 
We employ per-token asymmetric integer quantization for both Keys and Values, setting the quantization group size of RotateKV to 128 across all evaluations to demonstrate the accuracy and robustness of our method at relatively coarse quantization granularity.
The group size for grouped-head rotation is consistently set to 4 across all the models and tasks we tested. 
We employ FP8 to store the scale parameters and INT8 for the zero-points, since this approach does not affect the results but improves the overall compression rate.
The calibration process is highly efficient, taking less than five minutes on a single RTX 4090 GPU for the LLaMA-2-7B model using the WikiText-2 dataset. 
Additionally, the evaluation results demonstrate that calibration performed on WikiText-2 generalizes effectively to other datasets.
\subsection{Main Results and Analysis}
\subsubsection{Perplexity Evaluation.}
We use the original KV rotation and quantization scheme from QuaRot \cite{ashkboos2024quarot} as one of the baselines to highlight the improvements of our proposed adaptive rotations. 
Additionally, we include KVQuant \cite{hooper2024KVQuant} for PPL comparison, which leverages several techniques such as non-uniform quantization and per-vector dense-and-sparse quantization, demonstrating state-of-the-art PPL result.
To ensure a fair comparison, for QuaRot we only quantize the KV cache, and for KVQuant, we adopt the scheme that preserves 0.5\% FP16 outliers, which aligns the average bit-widths of our approach.
As shown in Table \ref{tab:PPL}, compared with the FP16 baseline, RotateKV shows a PPL degradation of 0.01 at 4-bit quantization, less than 0.1 at 3-bit, and less than 0.3 at 2-bit on LLaMA-2-13B.
Compared to QuaRot, which exhibits significant PPL degradation at 2-bit quantization, our method remains effective even at extremely low bit widths.
Compared to KVQuant, RotateKV consistently demonstrates a PPL improvement of around 0.1 at 2-bit quantization across all LLMs we tested. 
Notably, RotateKV uses the simpler integer quantization.
\subsubsection{GSM8K Evaluation.}
\begin{table}[t]
\vspace{-5mm}
\centering
\resizebox{0.75\columnwidth}{!}{%
\begin{tabular}{@{}c|cccc|c@{}}
\toprule
\multirow{2}{*}{Methods} & \multicolumn{4}{c|}{GSM8K (8-shot) Accuracy $\uparrow$} & \multirow{2}{*}{\begin{tabular}[c]{@{}c@{}}Average\\ Bits\end{tabular}} \\ \cmidrule(lr){2-5}
 & \begin{tabular}[c]{@{}c@{}}LLaMA\\ 2-7B\end{tabular} & \begin{tabular}[c]{@{}c@{}}LLaMA\\ 2-13B\end{tabular} & \begin{tabular}[c]{@{}c@{}}LLaMA\\ 3-8B\end{tabular} & \begin{tabular}[c]{@{}c@{}}Mistral\\ 7B\end{tabular} &  \\ \midrule
FP16 & 14.18 & 25.40 & 51.33 & 42.68 & 16 \\ \midrule
KIVI & 13.19 & 24.64 & 43.44 & 39.12 & 2.50 \\
GEAR & 12.96 & 22.74 & 44.88 & 39.50 & 4.03 \\
MiKV & 9.02 & 21.00 & 30.93 & 36.47 & 3.23 \\
ZipCache & 13.50 & 25.02 & 49.20 & 41.32 & 3.23 \\
\textbf{RotateKV} & \textbf{13.95} & \textbf{25.09} & \textbf{50.49} & \textbf{42.99} & \textbf{2.25} \\ \bottomrule
\end{tabular}%
}
\vspace{-2mm}
\caption{Evaluations on GSM8K with 8-shot CoT prompting. }
\label{tab:gsm8k}
\vspace{-3mm}
\end{table}
The GSM8K dataset \cite{cobbe2021gsm8k} is widely used to evaluate the arithmetic reasoning capabilities of LLMs. 
This task presents significant challenges and evaluations on it can effectively show the impact of compression methods on model performance.
As shown in the Table \ref{tab:gsm8k}, RotateKV maintains strong CoT reasoning capabilities, with less than 1.7\% performance degradation compared FP16 baseline. 
It outperforms existing methods even at lower average bit-widths, demonstrating the robustness of our approach. 
Notably, lower average bit-widths lead to higher compression ratio, enabling support for longer context lengths and larger batch sizes on the same GPU setup.
\subsubsection{LongBench and MileBench Evaluations.}
\begin{table*}[t]
\vspace{-5mm}
\centering
\resizebox{0.8\textwidth}{!}{%
\begin{tabular}{@{}c|c|cccc|c|cccc|c@{}}
\toprule
\multirow{3}{*}{Methods} & Benchmarks & \multicolumn{4}{c|}{LongBench} & Benchmarks & \multicolumn{4}{c|}{MileBench} & \multirow{3}{*}{Avg.} \\ \cmidrule(lr){2-11}
 & Tasks & QMSum & LCC & SAMSum & TriviaQA & Tasks & \begin{tabular}[c]{@{}c@{}}Scene\\ Transition\end{tabular} & \begin{tabular}[c]{@{}c@{}}Moving\\ Attribute\end{tabular} & \begin{tabular}[c]{@{}c@{}}Egocentric\\ Navigation\end{tabular} & \begin{tabular}[c]{@{}c@{}}Document\\ VQA\end{tabular} &  \\ \cmidrule(lr){2-11}
 & Metrics & Rouge-L & Edit Sim & Rouge-L & F1 & Metrics & Accuracy & Accuracy & Accuracy & Accuracy &  \\ \midrule
FP16 & \multirow{3}{*}{\begin{tabular}[c]{@{}c@{}}LLaMA-2\\ -7B\end{tabular}} & 21.1 & 66.7 & 41.3 & 87.4 & \multirow{3}{*}{\begin{tabular}[c]{@{}c@{}}LLaVA-v1.5\\ -7B\end{tabular}} & 73.0 & 47.5 & 33.5 & 45.5 & 52.0 \\
KIVI &  & 12.3 & 44.4 & 27.3 & 38.6 &  & 34.5 & 25.5 & 21.0 & 19.0 & 27.8 \\
\textbf{RotateKV} &  & \textbf{21.3} & \textbf{66.7} & \textbf{41.7} & \textbf{87.7} &  & \textbf{67.5} & \textbf{37.0} & \textbf{27.0} & \textbf{29.0} & \textbf{47.2} \\ \midrule
FP16 & \multirow{3}{*}{\begin{tabular}[c]{@{}c@{}}LLaMA-2\\ -13B\end{tabular}} & 21.3 & 66.6 & 43.5 & 87.4 & \multirow{3}{*}{\begin{tabular}[c]{@{}c@{}}LLaVA-v1.5\\ -13B\end{tabular}} & 70.5 & 50.0 & 26.5 & 46.0 & 51.5 \\
KIVI &  & 20.6 & 45.6 & 33.1 & 75.7 &  & 64.0 & 40.0 & 25.5 & 33.0 & 42.2 \\
\textbf{RotateKV} &  & \textbf{21.4} & \textbf{66.6} & \textbf{43.5} & \textbf{87.4} &  & \textbf{64.5} & \textbf{46.0} & \textbf{28.0} & \textbf{40.0} & \textbf{49.7} \\ \midrule
FP16 & \multirow{3}{*}{\begin{tabular}[c]{@{}c@{}}Mistral\\ -7B\end{tabular}} & 20.4 & 67.3 & 43.2 & 89.2 & \multirow{3}{*}{\begin{tabular}[c]{@{}c@{}}LLaVA-v1.6\\ -Mistral\\ -7B\end{tabular}} & 71.0 & 44.0 & 34.5 & 38.0 & 51.0 \\
KIVI &  & 19.0 & 58.3 & 41.0 & 81.6 &  & \textbf{67.5} & \textbf{47.0} & 28.0 & 40.0 & 47.8 \\
\textbf{RotateKV} &  & \textbf{20.5} & \textbf{67.4} & \textbf{42.9} & \textbf{89.2} &  & 64.5 & 46.0 & \textbf{29.5} & \textbf{42.0} & \textbf{50.3} \\ \bottomrule
\end{tabular}%
}
\vspace{-2mm}
\caption{Evaluations across a range of tasks from LongBench and MileBnech. 
The quantization is all set to 2-bit. 
For KIVI, the group size and the residual length are all set to 128. 
For RotateKV, the group size is set to 128.}
\vspace{-2mm}
\label{tab:longbench}
\end{table*}
\begin{figure*}[t]
\vspace{-2mm}
\centering
\begin{subfigure}[b]{0.48\textwidth}
    \centering
    \includegraphics[width=\textwidth]{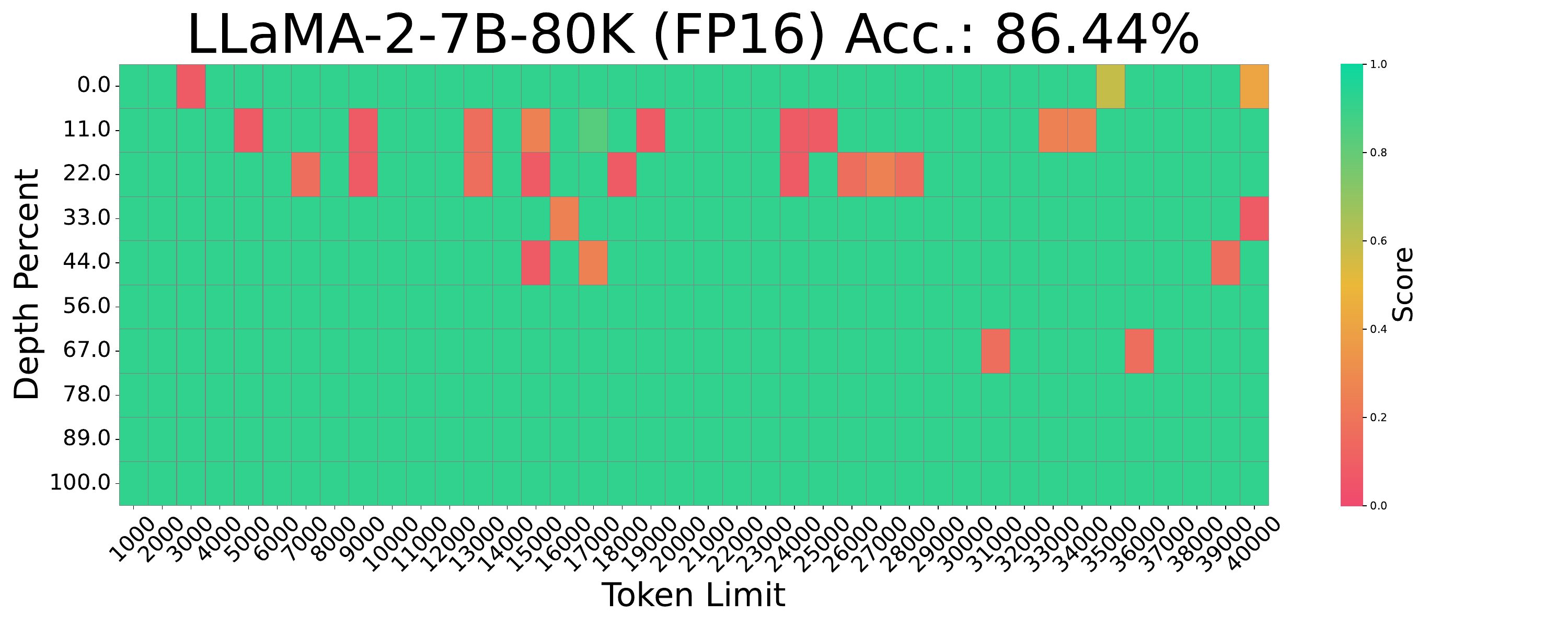}
    % \caption{NIAH test with FP16 KV cache.}
    \label{fig:niah_a}
\end{subfigure}\hfill
\begin{subfigure}[b]{0.48\textwidth}
    \centering
    \includegraphics[width=\textwidth]{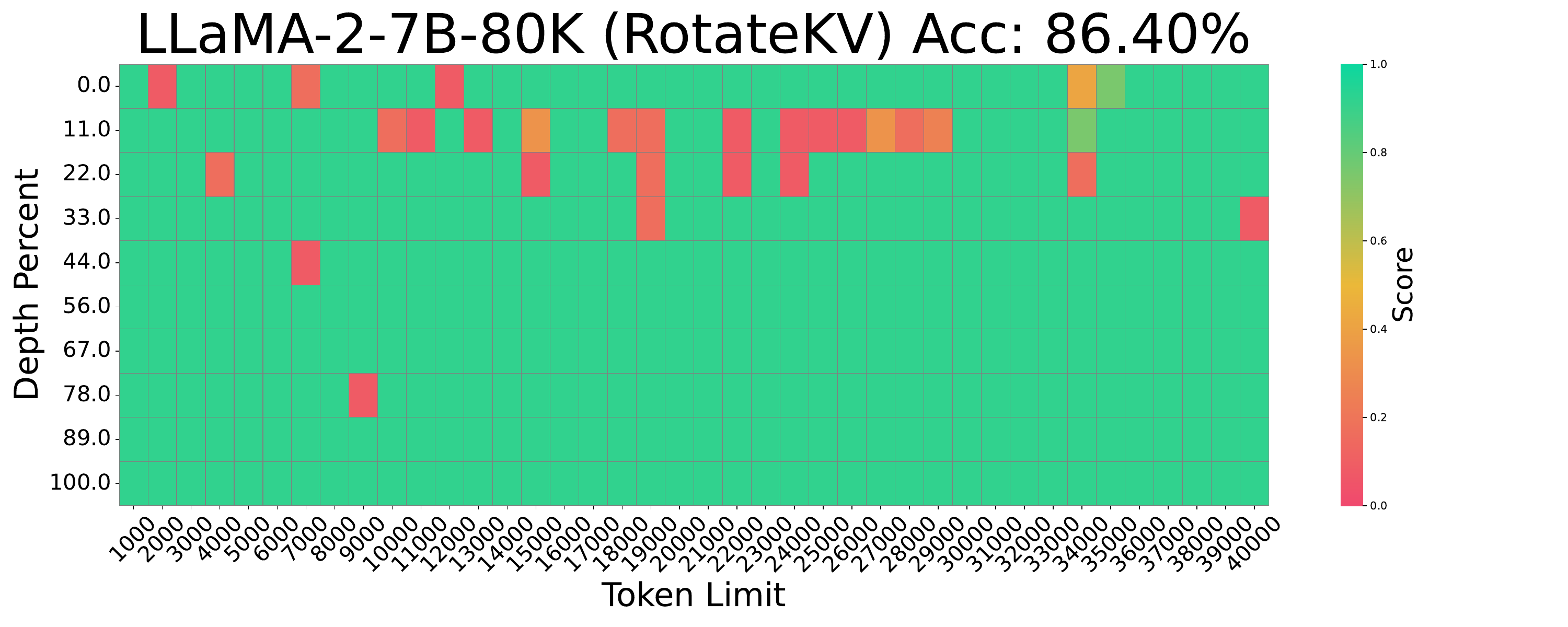}
        % \caption{NIAH test with RotateKV-2bit.}
    \label{fig:niah_b}
\end{subfigure}
\vspace{-7mm}
\caption{NIAH evaluation on the LLaMA-2-7B-80K model with a 40K context length.}
\label{fig:niah}
\vspace{-5mm}
\end{figure*}
To further demonstrate the robustness of RotateKV, we conduct experiments on eight long-context and multi-modal tasks selected from LongBench \cite{bai2023longbench} and MileBench \cite{song2024milebench}. 
For comparison, we include KIVI \cite{liu2024kivi}, which uses per-channel Key quantization and shows negligible performance degradation with a group size of 32.
As shown in Table \ref{tab:longbench}, compared to the FP16 baseline, KIVI with a 128 quantization group size \cite{liu2024kivi} exhibits significant performance degradation of 46.5\% on the LLaMA-2-7B and LLaVA-v1.5-7B models.
In contrast, our method demonstrates negligible average accuracy loss at the same quantization group size, with less than 1.4\% performance loss on Mistral-7B and LLaVA-v1.6-Mistral-7B.
\subsubsection{Needle-in-a-Haystack Evaluation.}
The NIAH task is designed to evaluate the ability to retrieve specific information within a large body of unrelated data. 
In our experiments, we utilize a context length of 40K, segmented into 40 intervals. 
Within each interval, the needle is positioned at 10 different depths of the context for evaluation. 
Figure \ref{fig:niah} demonstrates that RotateKV preserves retrieval capabilities with 2-bit KV cache quantization, underscoring the robustness of RotateKV in extremely long context scenarios.
\subsection{Ablation Study}
\begin{figure}[t]
\vspace{-1mm}
    \centering
    % First row: two images side by side using subfigures
    \begin{subfigure}[b]{0.45\linewidth}
        \centering
        \includegraphics[width=\linewidth]{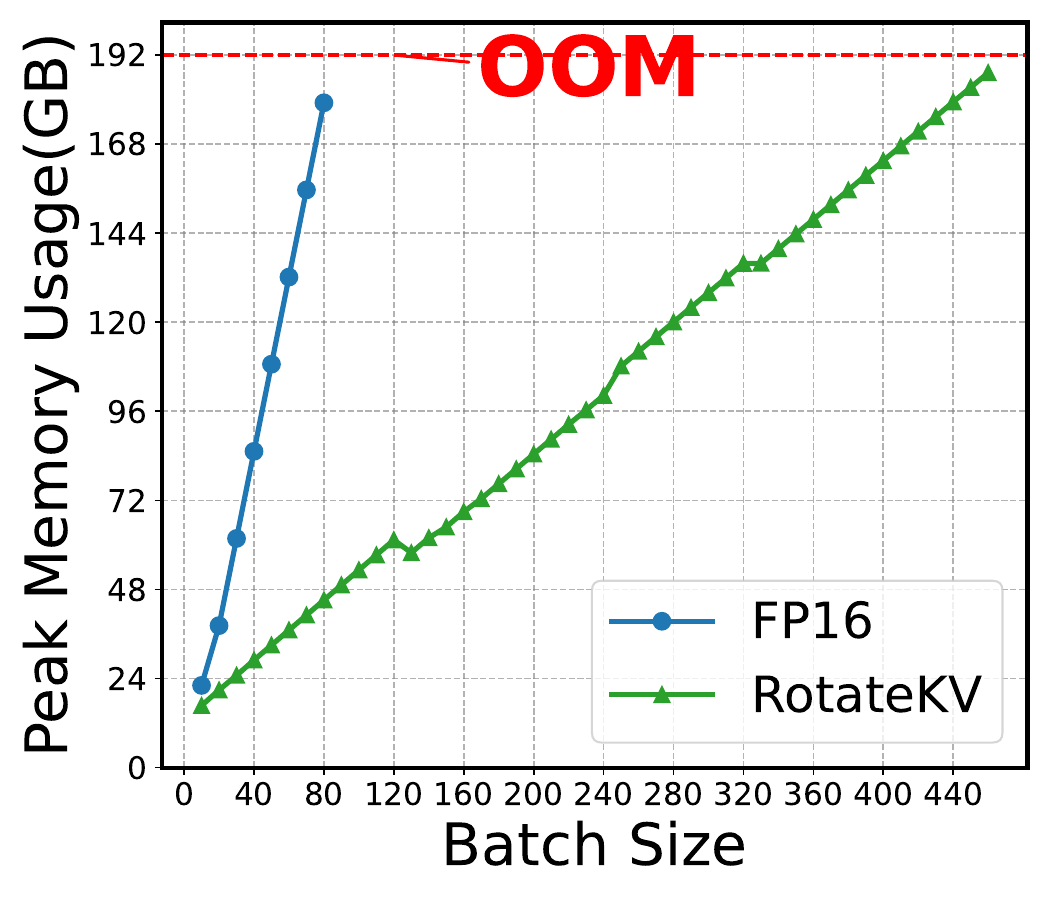}
        \vspace{-7mm}
        \caption{Memory evaluation.}
        \label{fig:memory}
    \end{subfigure} \hfill
    \begin{subfigure}[b]{0.45\linewidth}
        \centering
        \includegraphics[width=\linewidth]{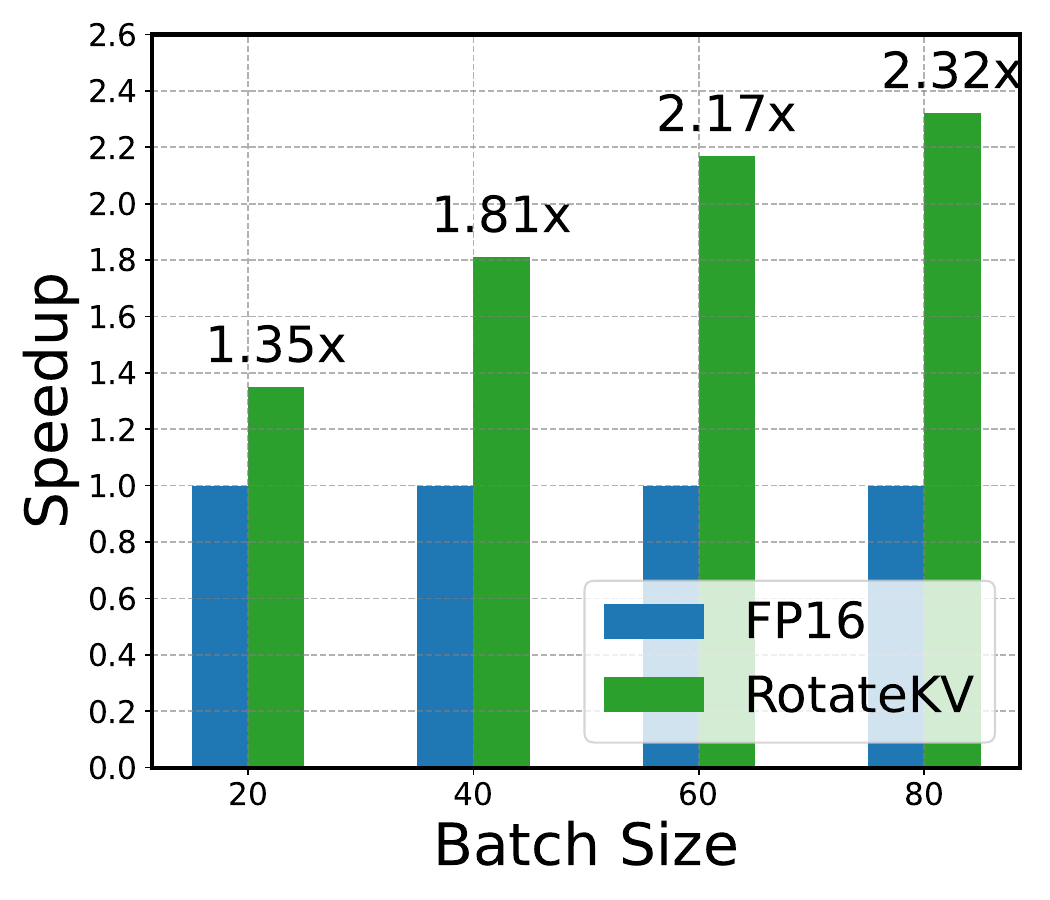}
        \vspace{-7mm}
        \caption{Speed evaluation.}
        \label{fig:speed}
    \end{subfigure}

    % Second row: a single image with a         \vspace{-5mm}
            \vspace{-3mm}
    \caption{Efficiency analysis of RotateKV. }
                \vspace{-5mm}
    \label{fig:efficiency}
\end{figure}
\subsubsection{Ablation Study of the Proposed Innovations}
Starting with the original rotation method from QuaRot \cite{ashkboos2024quarot}, we progressively incorporate the proposed innovations to evaluate their individual contributions. 
As shown in Table~\ref{tab:ablation}, the innovations introduced by RotateKV effectively mitigate PPL degradation.
\subsubsection{Ablation Study on Quantization Granularity}
To assess the impact of finer quantization granularity, we perform PPL evaluations with group sizes of 64 and 32. 
The results in Table \ref{tab:quantgrani} demonstrate that RotateKV continues to achieve better model performance preservation at smaller quantization group sizes. 
For instance, the PPL only drops by 0.18 on the LLaMA-2-13B model at 2-bit with a group size of 32.
\begin{table}[t]
\centering
\resizebox{0.6\columnwidth}{!}{%
\begin{tabular}{@{}lr@{}}
\toprule
LLaMA-2-13B & PPL $\downarrow$ \\ \midrule
FP16 & 4.57 \\
Original Rotations & 6.96 (2.39 $\uparrow$) \\ \midrule
+ Pre-RoPE Grouped-Head Key Rotation & 5.67 (1.29 $\downarrow$) \\
+ Attention-Sink-Aware Quantization & 5.52 (0.15 $\downarrow$) \\
+ Outlier-Aware Rotations & 4.84 (0.68 $\downarrow$) \\ \midrule
+ Scale (FP8) \& Zero-Point (INT8) & 4.84 \\ \bottomrule
\end{tabular}%
}
\vspace{-2mm}
\caption{Ablation study of the proposed innovations. 
We evaluate the PPL of 2-bit quantization with a sequence length of 4096 using the LLaMA-2-13B model on the WikiText-2 dataset.}
\vspace{-2mm}
\label{tab:ablation}
\end{table}
\begin{table}[t]
\vspace{-1mm}
\centering
\resizebox{0.4\columnwidth}{!}{%
\begin{tabular}{@{}c|cccc@{}}
\toprule
\multirow{2}{*}{\begin{tabular}[c]{@{}c@{}}Group\\ Size\end{tabular}} & \multicolumn{4}{c}{LLaMA-2-13B PPL $\downarrow$} \\ \cmidrule(l){2-5} 
 & \multicolumn{1}{c|}{16bit} & 4bit & 3bit & 2bit \\ \midrule
128 & \multicolumn{1}{c|}{\multirow{3}{*}{4.57}} & 4.58 & 4.64 & 4.84 \\
64 & \multicolumn{1}{c|}{} & 4.58 & 4.64 & 4.81 \\
32 & \multicolumn{1}{c|}{} & 4.58 & 4.66 & 4.75 \\ \bottomrule
\end{tabular}%
}
\vspace{-2mm}
\caption{Experiments on quantization granularity using the LLaMA-2-13B on the WikiText-2 dataset with a sequence length of 4096.}
\vspace{-4mm}
\label{tab:quantgrani}
\end{table}
\subsection{Efficiency Analysis}
In this section, we evaluate the efficiency of the current implementation. 
We use Triton \cite{tillet2019triton} for the quantization and dequantization kernels, along with an optimized CUDA kernel for FWHT, following \cite{dao2024fasthadamard}.
Evaluations are conducted on LLaMA-2-7B \cite{liu2024visual} with a 8-NVIDIA 4090D (24GB) setup, FlashAttention \cite{shah2024flashattention} enabled. 
The batch size is progressively increased with an input length of 500 tokens until an out-of-memory (OOM) error occurs. 
As shown in Figure \ref{fig:memory}, RotateKV reduces peak memory usage by 3.97× compared to the FP16 baseline and supports 5.75× larger batch sizes.
In terms of speed, RotateKV achieves a 2.32× speedup during the decoding phase, as shown in Figure \ref{fig:speed}.
Notably, the efficiency can be further improved through techniques like kernel fusion.
\vspace{-3mm}
\section{Conclusion}
In this paper, we explore the potential of rotation technique for 2-bit KV quantization. 
With the proposed innovations, RotateKV adaptively rotates the KV cache in an outlier-aware manner, demonstrating outstanding outliers control. 
Comprehensive evaluations show that RotateKV effectively preserves model performance even at high compression ratio, addressing the limitations of existing KV quantization methods and demonstrating state-of-the-art performance in both compression efficiency and accuracy.
Future work will focus on optimize RotateKV's implementation to further reduce the overhead associated with online operation during LLM inference.
% \section*{Acknowledgments}
% \bibliographystyle{named}
% \bibliography{ijcai25}

\end{document}